# Counting Belief Propagation

**Kristian Kersting**
Fraunhofer IAIS
Sankt Augustin, Germany

**Babak Ahmadi**
Fraunhofer IAIS
Sankt Augustin, Germany

**Sriraam Natarajan**
University of Wisconsin
Madison, USA

## Abstract

A major benefit of graphical models is that most knowledge is captured in the model structure. Many models, however, produce inference problems with a lot of symmetries not reflected in the graphical structure and hence not exploitable by efficient inference techniques such as belief propagation (BP). In this paper, we present a new and simple BP algorithm, called counting BP, that exploits such additional symmetries. Starting from a given factor graph, counting BP first constructs a compressed factor graph of clusternodes and clusterfactors, corresponding to sets of nodes and factors that are indistinguishable given the evidence. Then it runs a modified BP algorithm on the compressed graph that is equivalent to running BP on the original factor graph. Our experiments show that counting BP is applicable to a variety of important AI tasks such as (dynamic) relational models and boolean model counting, and that significant efficiency gains are obtainable, often by orders of magnitude.

## 1   Introduction

Message passing algorithms, in particular Belief Propagation (BP), have been very successful in efficiently computing interesting properties of probability distributions. Many graphical models, however, produce inference problems with a lot of symmetries not reflected in the graphical structure, and hence not exploited by BP. One of the most prominent examples are first-order and relational probabilistic models such as Markov logic networks [14]. Besides relational probabilistic models, however, there are also traditional, i.e., propositional probabilistic models that often produce inference problems with a lot of symmetries. In this work, we will demonstrate this for the classical model counting problem of computing the number of solutions of a given propositional formula. This problem vastly generalizes the NP-complete problem of propositional satisfiability, and hence is both highly useful and extremely expensive to solve in practice.

In this context, the present work makes two contributions. The key contribution is the introduction of *counting BP*, which is a BP approach that exploits additional symmetries and hence often scales much better than standard BP. Its underlying idea is rather simple: group together nodes and factors into clusternodes and clusterfeatures that are indistinguishable in terms of messages received and sent given the evidence. Exploiting this symmetry present in the probabilistic model makes it often possible to greatly compress the factor graph. More importantly, the compressed graph can be used to perform a modified BP yielding the same results as BP applied to the uncompressed factor graph. The second contribution is that we show that such symmetries are actually encountered in challenging AI tasks. Specifically, we apply counting BP to inference for dynamic relational probabilistic models and to model counting for Boolean formulas. As our experimental evaluation will show, in both application domains significant efficiency gains are obtainable, often by orders of magnitude.

We proceed as follows. We start of by discussing some related work. We briefly review standard BP in Section 3. In Section 4, we introduce *counting* BP. In Sections 5 and 6, we apply CBP to approximate inference for dynamic relational models and to model counting of Boolean formulas. Finally, we conclude and outline future research directions.

## 2   Related Work

The closest work to CBP is the recent work by Singla and Domingos [17]. Actually, an investigation of their approach was the seed that grew into our proposal we present in this paper. Singla and Domingos's *lifted first-order belief propagation* (LFOBP) builds upon [7] and also groups random variables, i.e., nodes that send and receive identical messages. CBP, however, differs from LFOBP in two important counts. First, CBP is conceptually easier than LFOBP. This is remarkable because efficient inference approaches for first-order and relational probabilistic models are typically rather complex. Second, LFOBP requires as input the specification of the probabilistic model in first-order logical format. Only nodes over the same predicate can be grouped together to form so-called supernodes. That means LFOBP coincides with standard BP for propositional MLNs, i.e., MLNs involving propositional variables only. The reason is that propositions are predicates with arity 0 so that the supernodes are singletons. Hence, no nodes and no features are grouped together. In contrast, CBP can directly be applied to any factor graph over finite random variables. In this sense, CBP is a generalization of LFOBP.

Sen *et al.* [16] recently presented another "clustered" inference approach based on bisimulation. Like CBP, which can also be viewed as running a bisimulation-like algorithm on the factor graph, Sen *et al.*'s approach also does not require a first-order logical specification. In contrast to CBP, it is guaranteed to converge but is also much more complex. Its efficiency in dynamic relational domains, in which variables easily become correlated over time by virtue of sharing common influences in the past, is unclear and its evaluation is an interesting future work.

Others such as Poole [13], Braz *et al.* [3, 4], and Milch *et al.* [9] have developed lifted versions of the variable elimination algorithm. They typically also employ a counting elimination operator that is equivalent to counting indistinguishable random variables and then summing them out immediately. These exact inference approaches are extremely complex, so far do not easily scale to realistic domains, and hence have only been applied to rather small artificial problems. Again, as for LFOBP, a first-order logical specification of the model is required.

## 3 Belief Propagation

Let $\mathbf{X} = (X_1, X_2, \ldots, X_n)$ be a set of $n$ discrete-valued random variables and let $x_i$ represent the possible realizations of random variable $X_i$. Graphical models compactly represent a joint distribution over $\mathbf{X}$ as a product of factors [12], i.e.,

$$P(\mathbf{X} = \mathbf{x}) = \frac{1}{Z} \prod_k f_k(\mathbf{x}_k) \ . \quad (1)$$

Here, each factor $f_k$ is a non-negative function of a subset of the variables $\mathbf{x}_k$, and $Z$ is a normalization constant. As long as $P(\mathbf{X} = \mathbf{x}) > 0$ for all joint configurations $\mathbf{x}$, the distribution can be equivalently represented as a log-linear model: $P(\mathbf{X} = \mathbf{x}) = \frac{1}{Z} \exp\left[\sum_i w_i \cdot g_i(\mathbf{x})\right]$, where the factors $g_i(x)$ are arbitrary functions of (a subset of) the configuration $\mathbf{x}$.

Graphical models can be represented as factor graphs. A factor graph, as shown in Fig 1, is a bipartite graph that expresses the factorization structure in Eq. (1). It has a variable node (denoted as a circle) for each variable $X_i$, a factor node (denoted as a square) for each $f_k$, with an edge connecting variable node $i$ to factor node $k$ if and only if $X_i$ is an argument of $f_k$. We will consider one factor $f_i(\mathbf{x}) = \exp\left[w_i \cdot g_i(\mathbf{x})\right]$ per feature $g_i(\mathbf{x})$, i.e., we will not aggregate factors over the same variables into a single factor.

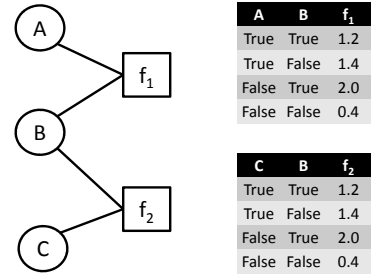

Figure 1: An example for a factor graph with associated potentials. Circles denote variables (binary in this case), squares denote factors.

An important inference task is to compute the conditional probability of variables given the values of some others, the evidence, by summing out the remaining variables. The belief propagation (BP) algorithm is an efficient way to solve this problem that is exact when the factor graph is a tree, but only approximate when the factor graph has cycles. One should note that the problem of computing marginal probability functions is in general hard (#P-complete).

We will now describe the BP algorithm in terms of operations on a factor graph. The computed marginal probability functions will be exact if the factor graph has no cycles, but the BP algorithm is still well-defined when the factor graph does have cycles. Although this loopy belief propagation has no guarantees of convergence or of giving the correct result, in practice it often does, and can be much more efficient than other methods [11].

To define the BP algorithm, we first introduce messages between variable nodes and their neighboring factor nodes and vice versa. The message from a variable $X$ to a factor $f$ is

$$\mu_{X \to f}(x) = \prod_{h \in \mathrm{nb}(X) \setminus \{f\}} \mu_{h \to X}(x) \quad (2)$$

where $\mathrm{nb}(X)$ is the set of factors $X$ appears in. The message from a factor to a variable is

$$\mu_{f \to X}(x) = \sum_{\neg \{X\}} \left( f(\mathbf{x}) \prod_{Y \in \mathrm{nb}(f) \setminus \{X\}} \mu_{Y \to f}(y) \right) \quad (3)$$

where $\mathrm{nb}(f)$ are the arguments of $f$, and the sum is over all of these except $X$, denoted as $\neg\{X\}$. The messages are usually initialized to 1.

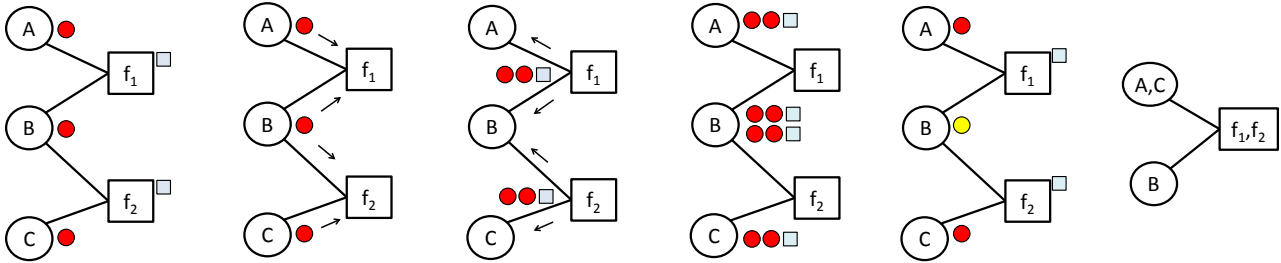

Figure 2: From left to right, the steps of CFG compressing the factor graph in Fig. 1 assuming no evidence. The shaded/colored small circles and squares denote the groups and signatures produced running CFG. On the right-hand side, the resulting compressed factor graph is shown. For details we refer to Section 4.

Now, the unnormalized belief of each variable $X_i$ can be computed from the equation

$$b_i(x_i) = \prod_{f \in \text{nb}(X_i)} \mu_{f \to X_i}(x_i) \qquad (4)$$

Evidence is incorporated by setting $f(\mathbf{x}) = 0$ for states $\mathbf{x}$ that are incompatible with it. Different schedules may be used for message-passing. We will touch upon this issue later again.

## 4 Counting Belief Propagation

Although already quite efficient, many graphical models produce factor graphs with a lot of symmetries not reflected in the graphical structure. Reconsider the factor graph in Fig. 1. The associated potentials are identical. In other words, although the factors involved are different on the surface, they actually share quite a lot of information. Standard BP cannot make use of this information. In contrast, *counting* BP – which we will introduce now – can make use of it and speed up inference by orders of magnitude.

Counting BP performs two steps: Given a factor graph $G$, it first computes a compressed factor graph $\mathfrak{G}$ and then runs a modified BP on $\mathfrak{G}$. We will now discuss each step in turn using fraktur letters such as $\mathfrak{G}$, $\mathfrak{X}$, and $\mathfrak{f}$ to denote compressed graphs, nodes, and factors.

**Step 1 – Compressing the Factor Graph:** Essentially, we simulate BP keeping track of which nodes and factors send the same messages, and group nodes and factors together correspondingly.

Let $G$ be a given factor graph with variable and factor nodes. Initially, all variable nodes fall into three groups (one or two of these may be empty), namely known true, known false, and unknown. For ease of explanation, we will represent the groups by colored/shaded circles, say, magents/white, green/gray, and red/black. All factor nodes with the same associated potentials also fall into one group represented by colored/shaded squares. For the factor graph in Fig. 1 the situation is depicted in Fig. 2. As shown on the left-hand side, assuming no evidence, all variable nodes are unknown, i.e., red/dark. Now, each variable node sends a message to its neighboring factor nodes saying "I am of color/shade red/black". A factor node sorts the incoming colors/shades into a vector according to the order the variables appear in its arguments. The last entry of the vector is the factor node's own color/shade, represented as light blue/gray square in Fig. 2. This color/shade signature is sent back to the neighboring variables nodes, essentially saying "I have communicated with these nodes". The variable nodes stack the incoming signatures together and, hence, form unique signatures of their one-step message history. Variable nodes with the same stacked signatures, i.e., message history can be grouped together. To indicate this, we assign a new color/shade to each group. In our running example, only variable node B changes its color/shade from red/black to yellow/gray. The factors are grouped in a similar fashion based on the incoming color/shade signatures of neighboring nodes. Finally, we iterate the process. As the effect of the evidence propagates through the factor graph, more groups are created. The process stops when no new colors/shades are created anymore.

The final compressed factor graph $\mathfrak{G}$ is constructed by grouping all nodes with the same color/shade into so-called *clusternodes* and all factors with the same color/shade signatures into so-called *clusterfactors*. In our case, variable nodes A, C and factor nodes $f_1, f_2$ are grouped together, see the right hand side of Fig. 2. Clusternodes (resp. clusterfactors) are sets of nodes (resp. factors) that send and receive the same messages at each step of carrying out BP on $G$. It is clear that they form a partition of the nodes in $G$.

Algorithm 1 summarizes our approach for computing the compressed factor graph $\mathfrak{G}$, on which we can run BP with minor modifications.

**Step 2 – BP on the Compressed Factor Graph:** Recall that the basic idea is to simulate BP carried out on $G$ on $\mathfrak{G}$. An edge from a clusterfactor $\mathfrak{f}$ to a cluster node $\mathfrak{X}_i$ in $\mathfrak{G}$ essentially represents multiple edges in $G$. Let $c(\mathfrak{f}, \mathfrak{X}_i)$ be

**Algorithm 1**: CFG – CompressFactorGraph

**Data**: A factor Graph $G$ with variable nodes $X$ and factors $f$, Evidence $E$
**Result**: Compressed Graph $\mathfrak{G}$ with clustervariable nodes $\mathfrak{X}$ and clusterfactor nodes $\mathfrak{f}$

1  Compute initial clusters of the $X_i$s w.r.t. $E$;
2  **repeat**
3     **foreach** *factor $f_k$* **do**
4        $signature_{f_k} = [\,]$;
5        **foreach** *node $X_i \in nb(f_k)$* **do**
6           $signature_{f_k}$.append($X_i.color$);
7        $signature_{f_k}$.append($f_k.color$);
8     Group together all $f_k$s having the same signature;
9     Assign each such cluster a unique color;
10    Set $f_k.color$ correspondingly for all $f_k$s;
11    **foreach** *node $X_i \in X, i = 1, \ldots, n$* **do**
12       $signature_{X_i} = [\,]$;
13       **foreach** *factor $f_k \in nb(X_i)$* **do**
14          $signature_{X_i}$.append($f_k.color$);
15       $signature_{X_i}$.append($X_i.color$);
16    Group together all $X_i$s having the same signature;
17    Assign each such cluster a unique color;
18    Set $X_i.color$ correspondingly for all $X_i$s;
19 **until** *grouping does not change* ;

the number of identical messages that would be sent from the factors in the clusterfactor $\mathfrak{f}$ to each node in the clusternode $\mathfrak{X}_i$ if BP was carried out on $G$. The message from a clustervariable $\mathfrak{X}$ to a clusterfactor $\mathfrak{f}$ is $\mu_{\mathfrak{X} \to \mathfrak{f}}(x) =$

$$\mu_{\mathfrak{f} \to \mathfrak{X}}(x)^{c(\mathfrak{f},\mathfrak{X})-1} \cdot \prod_{\mathfrak{h} \in \mathrm{nb}(\mathfrak{X}) \setminus \{\mathfrak{f}\}} \mu_{\mathfrak{h} \to \mathfrak{X}}(x)^{c(\mathfrak{h},\mathfrak{X})} \quad (5)$$

where $\mathrm{nb}(\mathfrak{X})$ now denotes the neighbor relation in the compressed factor graph $\mathfrak{G}$. The $c(\mathfrak{f}, \mathfrak{X}) - 1$ exponent reflects the fact that a clustervariable's message to a clusterfactor excludes the corresponding factor's message to the variable if BP was carried out on $G$.

The unnormalized belief of $\mathfrak{X}_i$, i.e., of any node $X$ in $\mathfrak{X}_i$ can be computed from the equation

$$b_i(x_i) = \prod_{\mathfrak{f} \in \mathrm{nb}(\mathfrak{X}_i)} \mu_{\mathfrak{f} \to \mathfrak{X}_i}(x_i)^{c(\mathfrak{f},\mathfrak{X})} \quad (6)$$

Evidence is incorporated by setting $\mathfrak{f}(\mathbf{x}) = 0$ for states $\mathbf{x}$ that are incompatible with it. Again, different schedules may be used for message-passing.

To conclude the section, the following theorem states the correctness of *counting* BP.

**Theorem 4.1.** *Given a factor graph $G$, there exists a unique minimal compressed $\mathfrak{G}$ factor graph, and algorithm CFG($G$) returns it. Running BP on $\mathfrak{G}$ using Eqs.* (5) *and* (6) *produces the same results as BP applied to $G$.*

The theorem generalizes the theorem of Singla and Domingos [17] but can essentially be proven along the same ways. The trick is to view potentials associated with factors as weighted clauses. Although very similar in spirit, counting BP has one important advantage: not only can it be applied to first-order and relational probabilistic models but also directly to traditional, i.e., propositional models such as Markov networks. We will demonstrate this by presenting two significant showcases for the application of counting BP: approximate inference for dynamic relational models and model counting of Boolean formulas.

## 5 Dynamic Relational Domains

Stochastic processes evolving over time are widespread. Traditionally, graphical models such as dynamic Bayesian networks [5] have been used to represent uncertain processes over time. DBNs represent the state of the world as a set of variables, and model the probabilistic dependencies of the variables within and between time steps. While DBNs can often yield compact representations, many real-world domains cannot be represented compactly using them: domains can contain multiple types of objects as well as multiple kinds of relations among them.

Formalisms that can represent objects and relations, as opposed to just random variables, have a long history in artificial intelligence. Recently, significant progress has been made in combining them with a principled treatment of uncertainty [6, 2]. First-order probabilistic models essentially combine graphical models with elements of first-order logic by defining template factors (such as Poole's parfactors [13]) that apply to whole sets of objects at once. A simple and powerful such language is Markov logic [14].

### 5.1 Dynamic Markov Logic Networks

A Markov logic network (MLN) (a social network example is shown in Table 1 (**Top**)) is a set of weighted first-order clauses. Together with a set of constants representing objects in the domain of interest, it defines a Markov network with one node per ground atom and one feature per ground clause. The weight of a feature is the weight of the first-order clause that originated it. The probability of a state $\mathbf{x}$ in such a network is given by $P(\mathbf{x}) = \frac{1}{Z} \exp\left[\sum_i w_i \cdot g_i(\mathbf{x})\right] = \frac{1}{Z} \prod_i f_i(\mathbf{x})$, where $w_i$ is the weight of the $i$th clause, $g_i = 1$ if the $i$th clause is true, $g_i = 0$ otherwise. Inference can be carried out by creating the ground network and applying belief propagation to it, but this can be extremely inefficient because the size of the ground network is in $\mathcal{O}(d^c)$, where $d$ is the number of domain objects and $c$ the highest clause arity.

In a stochastic logical process, the truth values of relations depend on the time step $t$. For instance, a smoker may quit smoking tomorrow. Therefore, we extend MLNs by allowing the modeling of time. The resulting framework is called dynamic MLNs (DMLNs).

| English | First-Order Logic | Weight |
|---|---|---|
| Most people do not smoke | $\neg \text{Smokes}(x)$ | 1.4 |
| Most people do not have cancer | $\neg \text{Cancer}(x)$ | 2.3 |
| Most people are not friends | $\neg \text{Friends}(x, y)$ | 4.6 |
| Smoking causes cancer | $\text{Smokes}(x) \Rightarrow \text{Cancer}(x)$ | 2.0 |
| Friends have similar smoking habits | $\text{Friends}(x, y) \Rightarrow (\text{Smokes}(x) <=> \text{Smokes}(y))$ | 2.0 |
| Apriori most people do not smoke | $\neg \text{Smokes}(x, 0)$ | 1.4 |
| Apriori most people do not have cancer | $\neg \text{Cancer}(x, 0)$ | 2.3 |
| Apriori most people are not friends | $\neg \text{Friends}(x, y, 0)$ | 4.6 |
| Smoking causes cancer | $\text{Smokes}(x, t) \Rightarrow \text{Cancer}(x, t)$ | 2.0 |
| Friends have similar smoking habits | $\text{Friends}(x, y, t) \Rightarrow (\text{Smokes}(x, t) <=> \text{Smokes}(y, t))$ | 2.0 |
| Most friends stay friends | $\text{Friends}(x, y, t) \Leftrightarrow \text{Friends}(x, y, \text{succ}(t))$ | 5.0 |
| Most smokers stay smokers | $\text{Smokes}(x, t) \Leftrightarrow \text{Smokes}(x, \text{succ}(t))$ | 5.0 |

Table 1: (**Top**) Example of a social network Markov logic network inspired by [17]. Free variables are implicitly universally quantified. (**Bottom**) Dynamic extension of the static social network model.

Specifically, we introduce *fluents*, a special form of predicates whose last argument is time. In this paper, we focus on discrete time processes, i.e., the time argument is non-negative integer valued. Furthermore, we assume a successor function $\text{succ}(t)$, which maps the integer $t$ to $t + 1$. There are two kinds of formulas: *intra-time* and *inter-time* ones. Intra-time formulas specify dependencies within a time slice and, hence, do not involve the succ function. In contrast, inter-time clauses involve the function succ. To enforce the Markov assumption, each term in the formula is restricted to at most one application of the succ function, i.e., terms such as $\text{succ}(\text{succ}(t))$ are disallowed. A dynamic MLN is now a set of weighted intra- and inter-time formulas. Given the domain constants, in particular the time range $0, \ldots, T_{\max}$ of interest, a DMLN induces a MLN and in turn a Markov network over time.

As an example consider the social network DMLN shown in Table 1 (**Bottom**). The first three clauses encode the initial distribution at $t = 0$. The next two clauses are intra-time clauses that talk about the relationships that exist within a single time-step. They say that smoking causes cancer and that friends have similar smoking habits. Of course, these are not hard clauses as with the case of first-order logic. The weights presented in the right column serve as soft-constraints for the clauses. The last two clauses are the inter-time clause and talk about friends and smoking habits persisting over time.

This model is similar to a dynamic Bayesian network except that it is undirected. Assume that there are two constants Anna and Bob. Let us say that Bob smokes at time 0 and he is friend with Anna. Then the ground Markov network will have a clique corresponding to the first two clauses for every time-step starting from 0. There will also be edges between Smokes(Bob) (correspondingly Anna) an between the Friends(Bob, Anna) for consecutive time-steps.

### 5.2 Lifted First-Order Factored Frontier

To perform inference, we could employ any known MLN inference algorithm. Unlike the case for static MLNs, however, we need approximation even for sparse models: Random variables easily become correlated over time by virtue of sharing common influences in the past.

Classical approaches to perform approximate inference in DBNs are the Boyen-Koller (BK) algorithm [1] and Murphy and Weiss's factored frontier (FF) algorithm [10]. Both approaches have been shown to be equivalent to one iteration of BP but on different graphs [10]. BK, however, involves exact inference, which for probabilistic logic models is extremely complex, so far does not scale to realistic domains, and hence has only been applied to rather small artificial problems. In contrast, FF is a more aggressive approximation. It is equivalent to (loopy) BP on the regular factor graph with a *forwards-backwards* message protocol: each node first sends message from "left" to "right" and then sends messages from "right" to "left". Hence, the basic idea of lifted first-order factored frontier (LFOFF) is to *plug in counting BP in place of BP* in FF.

### 5.3 Experimental Evaluation

We used the social network DMLN in Table 1 (**Bottom**). There were 20 people in the domain. For fractions $r \in \{0.0, 0.25, 0.5, 0.75, 1.0\}$ of people we randomly choose whether they smoke or not and who 5 of their friends are, and randomly assigned a time step to the information. Other friendship relations are still assumed to be unknown. $\text{Cancer}(x, t)$ is unknown for all persons x and all time steps. The "observed" people were randomly chosen. The query predicate was Cancer.

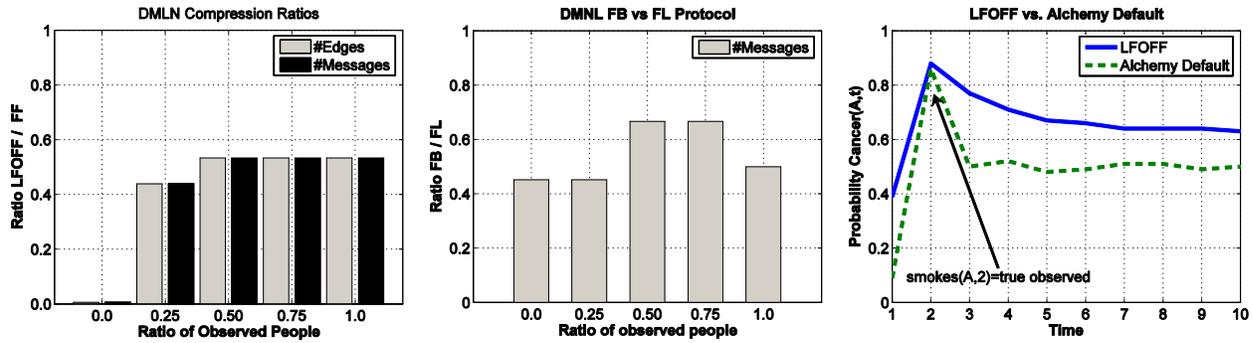

Figure 3: (**Left**) Ratios (LFOFF / FF) of number of edges and messages computed. The lower the value, the greater the speed-up when using LFOFF in place of FF. (**Middle**) Ratios (Forwards-Backwards / Flooding protocol) of number of messages computed. The lower the value, the greater the speed-up when using the FB protocol in place of the FL protocol. (**Right**) Probability estimates for cancer(A, t) over time.

In the first experiment, we investigated the compression ratio between standard FF and LFOFF for 10 and 15 time steps. Fig. 3 (**Left**) shows the results for 10 time steps. The results for 15 were similar and therefore omitted here. As one can see, the size of the factor graph as well as the number of messages sent is much smaller for LFOFF.

In the second experiment, we compared the "forwards-backwards" message protocol with the "flooding" protocol, the most widely used and generally best-performing method for static networks. Here, messages are passed from each variable to each corresponding factor and back at each step. Again, we considered 10 time steps. The results shown in Fig. 3 (**Middle**) clearly favor the FB protocol.

For a qualitative comparison, we finally computed the probability estimates for cancer(A, t) using LFOFF and MC-SAT, the default inference of the ALCHEMY system[1]. For MC-SAT, we used default parameters. There were four persons (A, B, C, and D) and we observed that A smokes at time step 2. All other relations where unobserved for all time steps. We expect that the probability of A having cancer has a peak at $t = 2$ smoothly fading out over time. Fig. 3 (**Right**) shows the results. In contrast to LFOFF, MC-SAT does not show the expected behaviour. The probabilities drop irrespective of the distance to the observation.

So far, the results clearly favor CBP over BP. A compression and thereby a speed-up, however, is not guaranteed. If there are no symmetries – such as the random 3-CNF in the next section – CBP essentially coincides with BP.

## 6 Model Counting

Model counting is the classical problem of computing the number of solutions of a given propositional formula. It vastly generalizes the NP-complete problem of propositional satisfiability, and hence is both highly useful and extremely expensive to solve in practice. Interesting applications include multi-agent reasoning, adversarial reasoning, and graph coloring, among others. In this section, we present a new approach to compute a probabilistic lower bound on the model count based on counting BP.

### 6.1 Counting using Belief Propagation

Our approach, called CBPCOUNT, is based on BPCOUNT for computing a probabilistic lower bound on the model count of a Boolean formula $F$, which was recently introduced by Kroc *et al.* [8]. The basic idea is to efficiently obtain a rough estimate of the "marginals" of propositional variables using belief propagation with damping. The marginal of variable $u$ in a set of satisfying assignments of a formula is the fraction of such assignments with $u = true$ and $u = false$ respectively. If this information is computed accurately enough, it is sufficient to recursively count the number of solutions of only one of "$F$ with $u = true$" and "$F$ with $u = false$", and scale the count up accordingly. Kroc *et al.* have empirically shown that BPCOUNT can provide good quality bounds in a fraction of the time compared to previous, sample-based methods.

BPCOUNT works as follows. A propositional variable $u$ is called *balanced* if it occurs equally often positively and negatively in all solutions of a given formula $F$. Now, BP-COUNT performs $t$ iterations, keeping track of the minimum count obtained over these iterations. In each iteration, (1) it computes the marginals of all variables running BP without evidence on the factor graph composed of the propositions in $F$ as variable nodes and the clauses in $F$ as factors, (2) identifies the most balanced variable $u$, uniformly randomly set $u$ to $true$ or $false$, (3) simplifies $F$ by performing any possible unit propagations, and (4) repeats the process. An exact counter such as CACHET [15] is called when the formula is sufficiently simplified. At this point, let $s$ denote the number of variables randomly set in this iteration before calling CACHET, and let $M_c$ be the model count of the residual formula returned by CACHET. The count of this iteration can be computed as $2^{s-\alpha} \times M_c$ where $\alpha$ is a "slack" factor pertaining to our probabilistic confidence in the bound. Here, $2^s$ can be seen as scaling up the residual count by a factor of 2 for every uniform random decision we made when fixing variables. After $t$ iterations, the minimum of the counts over all iterations is reported as lower bound for the model count of $F$. As Kroc *et al.* point out, the reported count is correct with probability $1-2^{-\alpha \cdot t}$.

---
[1] http://alchemy.cs.washington.edu/

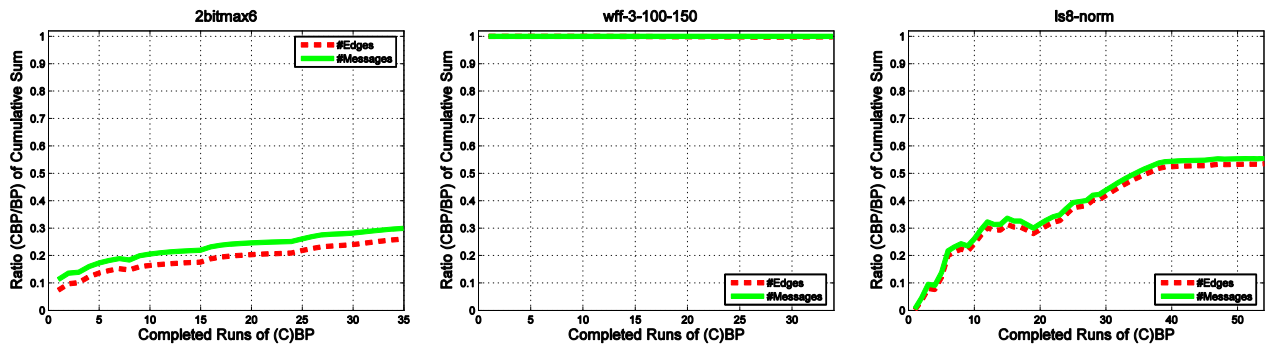

Figure 4: Ratios CBPCOUNT/BPCOUNT between 0.0 and 1.0 of the cummulative sum of edges computed respectively messages sent. A ratio of 1.0 means that CBP sends exactly as many messages as BP; a ratio of 0.5 that it sends half as many messages. (**Left**) `2bitmax_6`: Using CBP saved 88.7% of the messages BP sent in the first iteration of CBP-COUNT; in total, it saved 70.2% of the messages. (**Middle**) Random 3-CNF `wff-3-100-150`: No efficiency gain. The small difference in number of edges is due to a differently selected proposition due to tie breaking. (**Right**) `ls8-norm`: In the first iteration of CBPCOUNT, using CBP saved 99.4% of the messages BP sent. In total, this value dropped to 44.6%.

The basic idea of CBPCOUNT now is to *plug in counting BP in place of BP*. However, we have to be a little bit more cautious: propositional variables can appear at any position in the clauses. This makes high compression rates unlikely because, for each clusternode (set of propositional variables) and clusterfeature (set of clauses) combination, we carry a count for each position the clusternode appears in the clusterfeature. Fortunately, however, we deal with disjunctions only (assuming the formula $f$ is in CNF). Therefore, we can safely sort the message color signatures while compressing the factor graph. Reconsider the example from Fig. 2 and assume that the potentials associated with $f_1, f_2$ encode disjunctions. Indeed, assuming B to be the first argument of $f_1$ does not change the semantics of $f_1$. As our experimental results will show this can result in huge compression rates and large efficiency gains.

### 6.2 Experimental Evaluation

We have implemented (C)BPCOUNT based on SAMPLE-COUNT [2] using our own Python (C)BP implementation.

[2] www.cs.cornell.edu/~sabhar/software/samplecount/

We ran BPCOUNT and CBPCOUNT on the circuit synthesis problem `2bitmax_6` with damping factor 0.5 and convergence threshold $10^{-8}$. The formula has 192 variables, 766 clauses and a true count of $2.1 \times 10^{29}$. The resulting factor graph has 192 variable nodes, 766 factor nodes, and 1800 edges. The statistics of running (C)BPCount are shown in Fig. 4 (**Left**). As one can see, a significant improvement in efficiency is achieved when the marginal estimates are computed using CBP instead of BP: CBP reduces the messages sent by 88.7% when identifying the first, most balanced variable; in total, it reduces the number of messages sent by 70.2%. Both approaches yield the same lower bound of $5.8 \times 10^{28}$, which is in the same range as Kroc *et al.* report. Getting exactly the same lower bound was not possible because of the randomization inherent to BPCOUNT. Constructing the compressed graph took 9% of the total time of CBP. Overall, CBPCOUNT was about twice as fast as BPCOUNT, although our CBP implementation was not optimized.

Unfortunately, such a significant efficiency gain is not always obtainable. We ran BPCOUNT and CBPCOUNT on the random 3-CNF `wff-3-100-150`. The formula has 100 variables, 150 clauses and a true count of $1.8 \times 10^{21}$. Both approaches yield again the same lower bound, which is in the same range as Kroc *et al.* report. The statistics of running (C)BPCount are shown in Fig. 4 (**Middle**). CBP is not able to compress the factor graph at all. In turn, it does not gain any efficiency but actually produces a small overhead due to trying to compress the factor graph and to compute the counts.

In real-world domains, however, there is often a lot of redundancy. As a final experiment, we ran BPCOUNT and CBPCOUNT on the Latin square construction problem `ls8-norm`. The formula has 301 variables, 1601 clauses and a true count of $5.4 \times 10^{11}$. Again, we got similar estimates as Kroc *et al.*. The statistics of running (C)BPCount are shown in Figure 4 (**Right**). In the first iteration, CBP sent only $0.6\%$ of the number of messages BP sent. This corresponds to 162 times less many messages sent than BP.

To summarize, the experimental results show (a) CBP-COUNT can indeed speed-up BPCOUNT and (b) there are real world cases, in which CBP computes several orders of magnitude less many messages than BP.

## 7 Conclusions

The key contribution of this paper is the introduction of *counting BP*, a novel, scaleable belief propagation approach. CBP constructs a compressed factor graph of clustervariables and clusterfactors, corresponding to sets of nodes and factors that are indistinguishable given the evidence, and applies a modified belief propagation to this factor graph. It has been used to implement two novel algorithms for challenging AI tasks: a lifted factor frontier algorithm for approximate inference in dynamic Markov logic networks and an efficient approach for computing a lower bound on the model count for Boolean formulas. A number of experiments have shown that significant efficiency gains are obtainable when running counting BP instead of standard BP, often by orders of magnitude.

This work suggests several lines of future work such as approximate grouping of nodes and factors, developing generalized CBP variants, using CBP for (relational) learning, and applying it to real world domains. Most promising is to further explore the link to SAT-based techniques e.g. for efficient planning and first-order model counting.

**Acknowledgements** The authors would like to thank Pedro Domingos, Parag Singla, and the anonymous reviewers for their comments. Kristian Kersting and Babak Ahmadi gratefully acknowledge the support of the Fraunhofer ATTRACT fellowship STREAM. Sriraam Natarajan gratefully acknowledges support of the Defense Advanced Research Projects Agency under DARPA grants FA8650-06-C-7606 and HR0011-07-C-0060.